\documentclass[runningheads]{llncs}
\usepackage{graphicx}

\usepackage{tikz}
\usepackage{comment} 
\usepackage{amsmath,amssymb} %
\usepackage{color}

\usepackage{xcolor}
\usepackage{multirow}
\usepackage{textcomp}
\usepackage{diagbox}
\usepackage{hyperref}

\newcommand{\etal}{\textit{et al.}}
\newcommand{\ie}{\textit{i.e.}}

\begin{document}
\pagestyle{headings}
\mainmatter
\def\ECCVSubNumber{5445}  %
\title{Duality Diagram Similarity: \\ a generic framework for initialization selection \\ in task transfer learning} %

\titlerunning{Duality Diagram Similarity for Transfer Learning}

\author{Kshitij  Dwivedi\inst{1,3}\orcidID{0000-0001-6442-7140} \and
Jiahui Huang\inst{2}\orcidID{0000-0002-0389-1721} \and
Radoslaw Martin Cichy\inst{3}\orcidID{0000-0003-4190-6071} \and \\
Gemma Roig\inst{1}\orcidID{0000-0002-6439-8076} }

\authorrunning{Dwivedi et al.}
\institute{Department of Computer Science,  Goethe University Frankfurt, Germany \\
\email{kshitijdwivedi93@gmail.com, roig@cs.uni-frankfurt.de }\and
ISTD, Singapore University of Technology and Design, Singapore
\email{jiahui\_huang@sutd.edu.sg }\\ \and
Department of Education and Psychology, Free University Berlin, Germany\\
\email{rmcichy@zedat.fu-berlin.de}}
\maketitle

\begin{abstract}
In this paper, we tackle an open research question in transfer learning, which is selecting a model initialization to achieve high performance on a new task, given several pre-trained models. We propose a new highly efficient and accurate approach based on duality diagram similarity (DDS) between deep neural networks (DNNs). DDS is a generic framework to represent and compare data of different feature dimensions. We validate our approach on the Taskonomy dataset by measuring the correspondence between actual transfer learning performance rankings on 17 taskonomy tasks and predicted rankings. Computing DDS based ranking for $17\times17$ transfers requires less than 2 minutes and shows a high correlation ($0.86$) with actual transfer learning rankings, outperforming  state-of-the-art methods by a large margin ($10\%$) on the Taskonomy benchmark.   We also demonstrate the robustness of our model selection approach to a new task, namely Pascal VOC semantic segmentation. Additionally, we show that our method can be applied to select the best layer  locations within a DNN for transfer learning on 2D, 3D and semantic tasks on NYUv2 and Pascal VOC datasets.
\end{abstract}

\section{Introduction}

Deep Neural Networks (DNNs)  are state-of-the-art models to solve different visual tasks, \emph{c.f.}~\cite{krizhevsky2012imagenet,zamir2018taskonomy}. Yet, when the number of training examples with labeled data is small, the models tend to overfit during training. To tackle this issue, a common approach is to use transfer learning by selecting a pre-trained network on a large-scale dataset and use it as  initialization~\cite{ren2015faster,DeeperDP}.
\begin{figure}[t]
\label{figure1}
\begin{center}
   \includegraphics[width=1\linewidth]{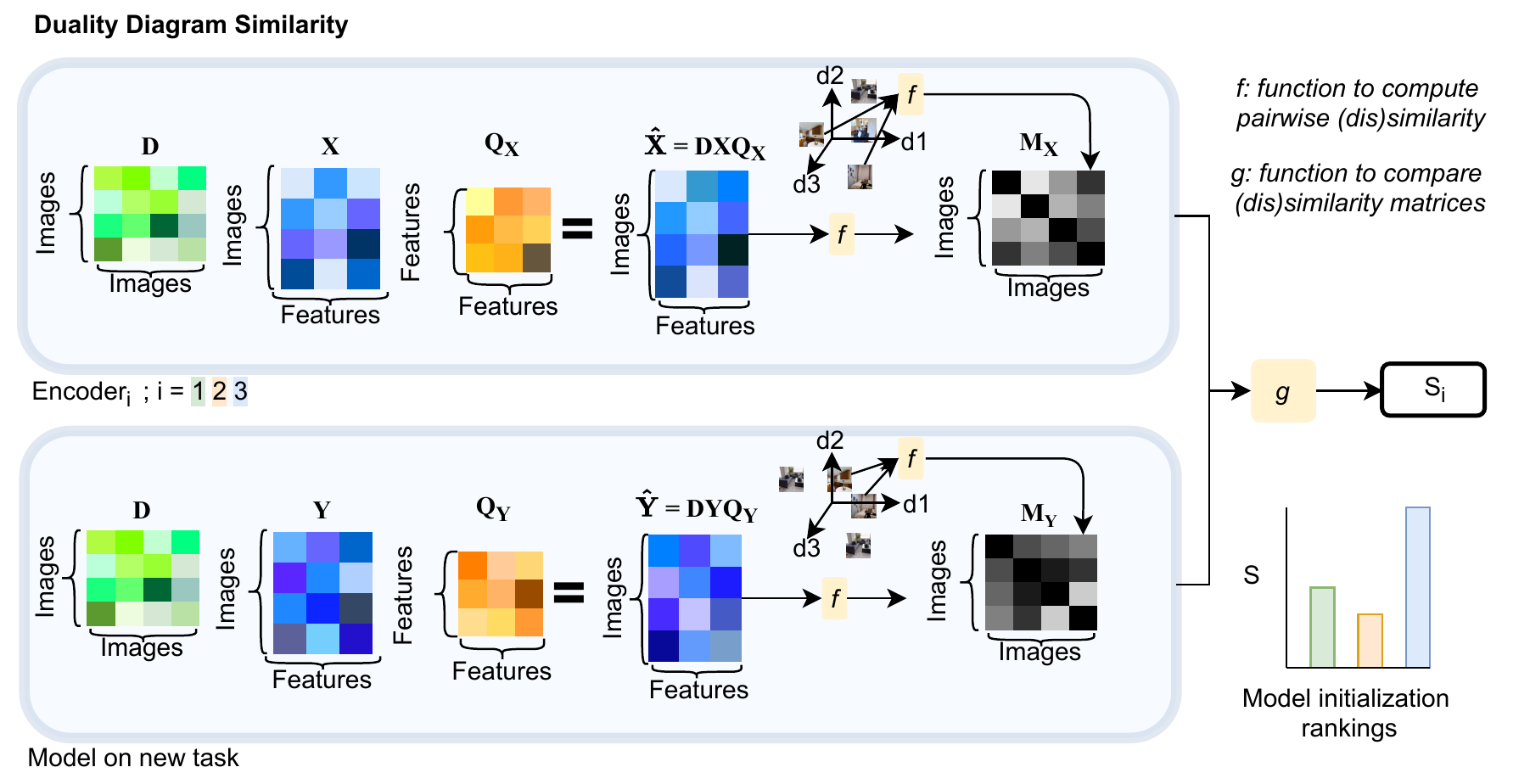}
\end{center}
   \caption{\emph{Duality Diagram Similarity (DDS):} We apply DDS to compare features of a set of initialization options (encoders) with features of a new task to get model initialization rankings to select the encoder initialization for learning a new task. The task feature for an image is obtained by doing a feedforward pass through a model trained on that task. $\mathbf{D}$ is the matrix that weights the images, $\mathbf{X}$ ($\mathbf{Y}$) is the matrix that stores the features from the encoder for all images, $\mathbf{Q_X}$ ($\mathbf{Q_Y}$) is a matrix that stores relations between features dimensions, $\mathbf{M_X}$ ($\mathbf{M_Y}$) contains the pairwise (dis)similarity distances between images, and $S_i$ is the score for the ranking.}%
   \label{fig:overview}
\end{figure}
But how does one choose the model initialization that yields the highest accuracy performance when learning a new task?

Nowadays, there are a plethora of online available pre-trained models on different tasks. However, there are only a few methods \cite{Dwivedi_RSA_19,song2019deep} that automatically assist in selecting an initialization given a large set of options. Due to lack of a standard benchmark with standard evaluation metrics, comparing and building upon these methods is not trivial. Recently, Dwivedi and Roig \cite{Dwivedi_RSA_19} and Song \etal~\cite{song2019deep} used the transfer learning performance on the Taskonomy dataset \cite{zamir2018taskonomy} as groundtruth to develop methods for model selection. Both aforementioned methods for model selection are efficient compared to the bruteforce approach of obtaining transfer performance from all the models and selecting the best one. Yet, they used different metrics to evaluate against the groundtruth, and hence, they are not comparable in terms of accuracy. Although different, both of them used metrics that evaluate how many models in top-K ranked model initializations according to transfer learning performance were present in the top-K ranked models obtained using their method. We argue that such a metric doesn't provide a complete picture as it ignores the ranking within the top-K models as well as the ranking of models not in the top-K.  

In this work, we first introduce a benchmark with a standard evaluation metric using Taskonomy~\cite{zamir2018taskonomy} transfer learning dataset to compare different model initialization selection methods. We use Spearman's correlation between the rankings of different initialization options according to transfer learning performance and the rankings based on a model initialization selection method as our metric for comparison. We argue that our proposed benchmark will facilitate the comparison of existing and new works on model selection for transfer learning.  We then introduce a duality diagram~\cite{escoufier1987duality,holmes2008multivariate,de2011duality} based generic framework to compare DNN features which we refer to as duality diagram similarity (DDS). Duality diagram expresses the data taking into account the contribution of individual observations and individual feature dimensions, and the interdependence between observations as well as feature dimensions (see Fig.~\ref{fig:overview}). Due to its generic nature, it can be shown that recently introduced similarity functions \cite{Dwivedi_RSA_19,kornblith2019better} for comparing DNN features are special cases of the  general DDS framework.

We find that model initialization rankings using DDS show very high correlation (\textgreater 0.84) with transfer learning rankings on Taskonomy tasks and outperform state-of-the-art methods~\cite{Dwivedi_RSA_19,song2019deep} by a 10\% margin. We also demonstrate the reliability of our method on a new dataset and task (PASCAL VOC semantic segmentation) in the experiments section.

Previous works~\cite{yosinski2014transferable,misra2016cross} have shown the importance of selecting which layer in the network to transfer from. In this paper, we also explore if the proposed method could be used to interpret representations at different depths in a pre-trained model, and hence, it could be used to select from which layer the initialization of the model should be taken to maximize transfer learning performance. We first show that the representation at different blocks of pre-trained ResNet \cite{ResNet} model on ImageNet~\cite{deng2009imagenet} varies from 2D in block1, to 3D in block 3 and semantic in block 4. These observations suggest that representation at different depths of the network is suitable for transferring to different tasks. Tranfer learning experiments using different  blocks in a ResNet-50 trained on ImageNet  as initialization for   2D, 3D, and semantic tasks on both, NYUv2~\cite{NYUV2} and Pascal VOC~\cite{PascalVOC} datasets, reveal that it is indeed the case.

\section{Related Works}

Our work relies on comparing DNN features to select pre-trained models as initialization for transfer learning. Here, we first briefly discuss related literature in transfer learning, and then, different methods to compare DNN features.  

\subsubsection{Transfer Learning.}
In transfer learning \cite{pan2009survey}  the   representations from a source tasks are re-used and adapted to a new target task. While transfer learning in general may refer to task transfer\cite{ren2015faster,DeeperDP,xie2015holistically,zamir2018taskonomy}, or domain adaptation\cite{rozantsev2018beyond,tzeng2017adversarial}, in this work we focus specifically on task transfer learning. Razavian \etal~\cite{sharif2014cnn} showed that features extracted from Overfeat~\cite{sermanet2013overfeat} network trained on ImageNet~\cite{deng2009imagenet} dataset can serve as a generic image representation to tackle a wide variety of recognition tasks. ImageNet pre-trained models also have been used to transfer to a diverse range of other vision related tasks~\cite{ren2015faster,DeepLabV3,DeeperDP,xie2015holistically}. Other works~\cite{kornblith2019better,huh2016makes} have investigated why ImageNet trained models are good for transfer learning. In contrast, we are interested in improving the transfer performance by finding a better initialization that is more related to the target task.  

Azizpour~\etal~\cite{azizpour2015factors} investigated different transferability factors. They empirically verified that the  effectiveness of a factor is highly correlated with the distance between the source and target task distance obtained with a predefined categorical task grouping. Zamir~\etal~\cite{zamir2018taskonomy} showed in a fully computational manner that initialization matters in transfer learning. Based on transfer performance they obtained underlying task structure that showed clusters of 2D, 3D, and semantic tasks. They introduced the Taskonomy dataset~\cite{zamir2018taskonomy}, which provides pre-trained models on over 20 single image tasks with transfer learning performance on each of these tasks with every pre-trained model trained on other tasks as the initialization, and thus, providing groundtruth for a large number of transfers. Recent works~\cite{Dwivedi_RSA_19,song2019deep} have used the Taskonomy transfer performance as groundtruth to evaluate methods of estimating task transferabilities. Those works use different evaluation metrics, which makes the comparison between those methods  difficult. Following those works, we use Taskonomy transfer performance as a benchmark, and propose a unified evaluation framework to facilitate comparison between existing and future methods.

Yosinski~\etal~\cite{yosinski2014transferable} explored transferability at different layers of a pre-trained network, and Zhuo \etal~\cite{Zhuo2017DeepUC} showed the importance of focusing on convolutional layers of the model in domain adaptation. We also investigate if the similarity between DNN representations can be applied to both model and layer selection for transfer learning, which indeed is the case, as we show in the results section.

\subsubsection{Similarity Measures for Transfer Learning Performance.}
Our approach is built under the assumption that the higher the similarity between representations is, the higher will be the transfer learning performance. Some previous works used similarity measures to understand the properties of DNNs. Raghu \etal~\cite{raghu2017svcca} proposed affine transform invariant measure called Singular Vector Canonical Correlation Analysis (SVCCA) to compare two representations. They applied SVCCA to probe the learning dynamics of neural networks. Kornblith \etal~\cite{kornblith2019similarity} introduced centered kernel alignment (CKA) that shows high reliability in identifying correspondences between representations in networks trained using different initializations. However, in the above works, the relation between similarity measures and transfer learning was not explored.

Dwivedi and Roig \cite{Dwivedi_RSA_19} showed that Representational Similarity Analysis (RSA) can be used to compare DNN representations. They argued that using the model parameters from a model that has a similar representation to the new task's representation as initialization, should give higher transfer learning performance compared to an initialization from a model with a lower similarity score. Recently, Song \etal~\cite{song2019deep} used attribution maps~\cite{simonyan2014very,bach2015pixel,shrikumar2016not} to compare two models and showed that it also reflects transfer learning performance. Our work goes beyond the aforementioned ones. Besides proposing an evaluation metric to set up a benchmark for comparison of these methods, we introduce a general framework using duality diagrams for similarity measures. We show that similarity measures, such as RSA and CKA, can be posed as a particular case in our general formulation. It also allows  to use other more powerful similarities that are more highly correlated to transfer learning performance. 

There is evidence in the deep learning literature, that normalization plays a crucial role. For instance, batch normalization allows training of deeper networks~\cite{ioffe2015batch}, efficient domain adaptation~\cite{Li2016RevisitingBN,Balaji2019NormalizedWF} and parameter sharing across multiple domains~\cite{rebuffi2018efficient}. Instance normalization improves the generated image quality in fast stylization~\cite{ulyanov2016instance,huang2017arbitrary}, and group normalization stabilizes small batch training~\cite{wu2018group}. In our DDS generic framework, it is straightforward to incorporate feature normalization.  Thus, we further take into account the normalization of features before assessing the similarity between two DNN features and compare it to transfer learning performance.

\section{Duality Diagram Similarity (DDS)}

The term duality diagram was introduced by Escoufier \cite{escoufier1987duality} to derive a general formula of Principal Component Analysis that takes into account change of scale, variables, weighing of feature dimensions and elimination of dependence between samples. With similar motivation, we investigate the application of duality diagrams in comparing two DNNs.  Let $\mathbf{X} \in \mathbb{R}^{n\times d_1}$ refer to a matrix of features with dimensionality $d_1$ obtained from feedforwarding $n$ images through a DNN. The duality diagram of matrix $\mathbf{X} \in \mathbb{R}^{n\times d_1}$ is a triplet ($\mathbf{X}$,$\mathbf{Q}$,$\mathbf{D}$) consisting of a matrix $\mathbf{Q} \in \mathbb{R}^{d_1\times d_1}$ that quantifies dependencies between the individual feature dimensions, and  a matrix $\mathbf{D} \in \mathbb{R}^{n\times n}$ that assigns weights on the observations, \ie, images in our case.  Hence, a DNN representation for a set of $n$ examples can be expressed by its duality diagram. By comparing duality diagrams of two DNNs we can obtain a similarity score. We denote the two DNN duality diagrams as ($\mathbf{X}$,$\mathbf{Q_X}$,$\mathbf{D}$) and ($\mathbf{Y}$,$\mathbf{Q_Y}$,$\mathbf{D}$),  in which the subindices in the matrix $\mathbf{Q}$ denote that they are computed from the set of features and images in $\mathbf{X}$ and $\mathbf{Y}$.

\begin{table*}[!t]

\centering

{
\begin{tabular}{c|c|c|c}
 
\emph{Distances} & Pearson's: & Euclidean: & cosine: \\

  & $1-\frac {({\textbf{x}_i-\overline{\textbf{x}_i})} \cdot {(\textbf{x}_j-\overline{\textbf{x}_j}})}{||{\textbf{x}_i-\overline{\textbf{x}_i}}|| \cdot ||{\textbf{x}_j-\overline{\textbf{x}_j}}||}$
  & $\sqrt{\textbf{x}_i^T.\textbf{x}_i+\textbf{x}_j^T.\textbf{x}_j-2*\textbf{x}_i^T.\textbf{x}_j}$ 
  & $1-\frac {{\textbf{x}_i} \cdot {\textbf{x}_j}}{||{\textbf{x}_i}|| \cdot ||{\textbf{x}_j}||}$  \\
   \hline
 
\emph{Kernels} & linear: & Laplacian: & RBF: 
\\ 

  & $\textbf{x}_i^T\textbf{x}_j$ 
  & $\exp(-{\gamma}_{1}||\textbf{x}_i - \textbf{x}_j||_1)$ 
  & $\exp(-{\gamma}_{2}||\textbf{x}_i - \textbf{x}_j||^2)$  \\ 
\end{tabular}
}
\caption{\emph{Distance and Kernel functions used in DDS.} Notation:   $\textbf{x}_i \in \mathbb{R}^{d1}$ and $\textbf{x}_j \in \mathbb{R}^{d1}$ refer to the features corresponding to ${i}^{th}$ and ${j}^{th}$ image (${i}^{th}$ and ${j}^{th}$ row of feature matrix $\mathbf{X}$), respectively. Here, ${\gamma}_{1}$ and ${\gamma}_{2}$ refer to the bandwidth of Laplacian and RBF kernel.}%
\label{table1}
\end{table*}

To compare two duality diagrams, Robert and Escoufier \cite{robert1976unifying} introduced the RV coefficient. The motivation behind RV coefficient was to map $n$ observations of $\mathbf{X}$ in the ${d_1}$-dimensional space and $\mathbf{Y}$ in the ${d_2}$-dimensional space. Then, the similarity between $\mathbf{X}$ and $\mathbf{Y}$ can be assessed by comparing the pattern of obtained maps
or, equivalently, by comparing the set of distances between all pairwise observations of both maps. To estimate the distances between pairwise observation, Robert and Escoufier \cite{robert1976unifying} used dot product and compared two (dis)similarity matrices using the cosine distance.  

In a nutshell, to compare two sets of DNN features $\mathbf{X}$ and $\mathbf{Y}$, we require three steps (Fig.~\ref{fig:overview}): first, transforming the data using $\mathbf{Q_X}$ and $\mathbf{D}$ to $\mathbf{\hat{X}}$, using $\mathbf{\hat{X}} = \mathbf{D}\mathbf{X}\mathbf{Q_X}$, and $\mathbf{\hat{Y}}$ with $\mathbf{\hat{Y}} = \mathbf{D}\mathbf{Y}\mathbf{Q_Y}$. Second, using a function, which we denote as $f$, to measure (dis)similarity between each pair of data points to generate pairwise distance maps. Let $\mathbf{M_X}$ be the matrix that stores the (dis)similarity between pairwise distance maps for $\mathbf{\hat{X}}$, also referred to as representational (dis)similarity matrices. It is computed as $\mathbf{M_X}(i,j)=f(\mathbf{\hat{X}}(i,:),\mathbf{\hat{X}}(j,:))$, in which $i$ and $j$ denote the indices of the matrices. Analogously, $\mathbf{M_Y}$ is the matrix that stores the (dis)similarity between pairwise distance maps of $\mathbf{\hat{Y}}$.  Third, a function $g$ to compare $\mathbf{M_X}$ and $\mathbf{M_Y}$ to obtain a final similarity score, denoted as $S$, and computed as $S=g(\mathbf{M_X},\mathbf{M_Y})$ is applied. The above formulation using duality diagrams provides a general formulation that allows us to investigate empirically which combination of $\mathbf{Q}$ ,$\mathbf{D}$, $f$ and $g$ is suitable for a given application, which in our case is estimating transferability rankings to select the best model (or layer in a model) to transfer given a new dataset and/or task. 

\begin{figure}[t!]
\begin{center}
   \includegraphics[width=1\linewidth]{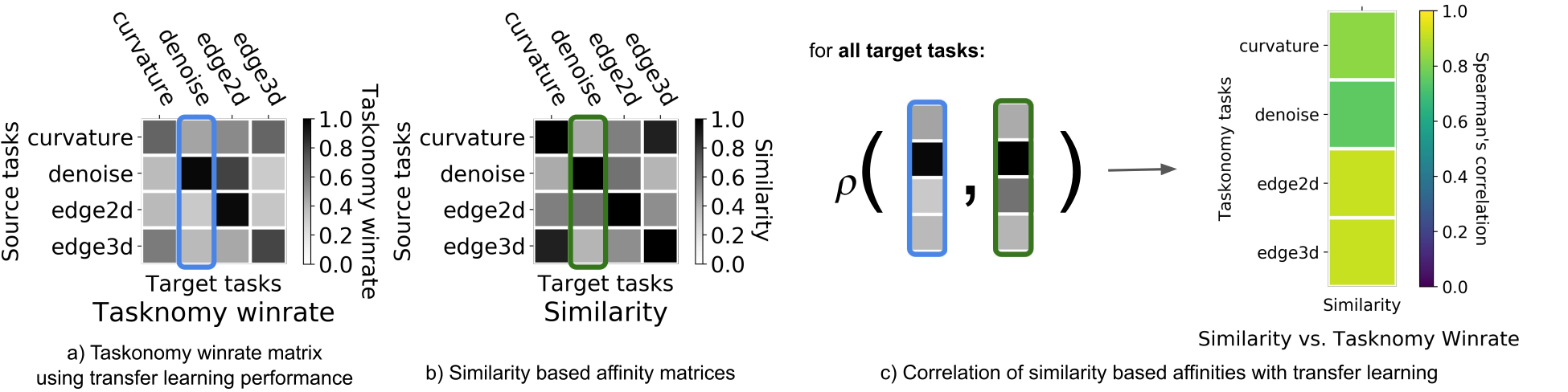}
\end{center}
   \caption{\emph{Transfer learning vs. similarity measures.} We consider a) Taskonomy winrate matrix, b) an affinity matrix obtained by measuring similarity between DNNs trained on different tasks. c) the Spearman's correlation (denoted by $\rho$) between the columns of two matrices. The resulting vector shows the correlation of the similarity based rankings with transfer learning performance based rankings for 4 Taskonomy tasks. Here we illustrate the results using DDS ($f=Laplacian$), and the procedure remains the same using any similarity measure.}%
\label{figure2}
\end{figure}

Interestingly, using the above DDS framework, we can easily show that recently used similarity measures, e.g., CKA and RSA, can be formulated as special cases of DDS. For RSA~\cite{Dwivedi_RSA_19},  $\mathbf{Q}$ is an identity matrix, $\mathbf{I} \in \mathbb{R}^{d_1\times d_1}$, and  $\mathbf{D}$ is a centering matrix, \ie, $\mathbf{C} = \mathbf{I_n}-\frac{1}{n}\mathbf{1_n}$. $f$ is Pearson's distance and $g$ is Spearman's correlation between lower/upper triangular part of $\mathbf{M_X}$ and $\mathbf{M_Y}$. For CKA~\cite{kornblith2019better}, $\mathbf{Q}$ and  $\mathbf{D}$ are identity matrices $\mathbf{I} \in \mathbb{R}^{d_1\times d_1}$ and $\mathbf{I} \in \mathbb{R}^{n\times n}$ respectively, $f$ used is linear or RBF kernel and $g$ is cosine distance between unbiased centered (dis)similarity matrices. In the supplementary section S1, we derive RSA and CKA as particular cases of the DDS framework. 

In this work, we focus on exploring different instantiations of $\mathbf{Q}$, $\mathbf{D}$, $f$ and $g$ from our DDS framework that are most suitable for estimating transfer learning performance. We consider different formulations of $\mathbf{Q}$ and $\mathbf{D}$, resulting in z-scoring, batch normalization, instance normalization, layer normalization and group normalization (details in Supplementary S2). %
For function $f$ we explore cosine, Euclidean, and Pearson's distance, as well as kernel based similarities, namely linear, RBF, and Laplacian. Mathematical equations for all functions are in Table \ref{table1}. For function $g$, we consider  Pearson's correlation to compare (dis)similarity matrices with and without unbiased centering~\cite{szekely2014partial,kornblith2019better}.%

\section{Our Approach}

\subsection{Which DDS combination ($\mathbf{Q}$, $\mathbf{D}$, $f$ ,$g$) best predicts transferability?}

After having defined the general formulation for using similarity measures for transfer learning, we can instantiate each of the parameters ($\mathbf{Q}$, $\mathbf{D}$, $f$ and $g$) to obtain different similarity measures. To evaluate which combination of $\mathbf{Q}$, $\mathbf{D}$, $f$ and $g$ best predicts transferability and compare it to state-of-the-art methods, we consider transfer learning performance based winrate matrix (Fig. \ref{figure2}a) and affinity matrix proposed in Taskonomy dataset \cite{zamir2018taskonomy}, as a transferability benchmark. The affinity matrix is calculated by using actual transfer learning performance on the target task given multiple source models pre-trained on different tasks.  The winrate matrix is calculated using a pairwise competition between all feasible sources for transferring to a target task. Both these matrices represent transfer learning performance obtained by bruteforce, and hence, can be considered as an upper bound for benchmarking transferability. We use the Taskonomy dataset as a benchmark as it consists of pre-trained models on over 20 single image tasks with transfer learning performance on each of these tasks with every pre-trained model trained on other tasks as the initialization, thus, providing groundtruth for a large number of task transfers. 

We use DDS to quantify the similarity between two models trained on different Taskonomy tasks and use that value to compute the DDS based affinity matrix (Fig. \ref{figure2}b). A column vector corresponding to a specific task in the Taskonomy affinity matrix shows the transfer learning performance on the target task when different source tasks were used for initialization. To evaluate how well a DDS based affinity matrix represents transferability, we calculate the Spearman's correlation between columns of the Taskonomy winrate/affinity matrix and DDS based affinity matrix. Using the rank-based Spearman's correlation for comparison between two rankings allows comparing the source tasks ranking on the basis of transfer learning performance with DDS based ranking. The resulting vector (Fig. \ref{figure2}c) represents the per task correlation of DDS with transferability.

We further evaluate if the best combination(s) we obtained from the above proposed evaluation benchmark using Taskonomy are robust to a new dataset and task. For this, we consider a new task, Pascal VOC semantic segmentation, following \cite{Dwivedi_RSA_19}. For the benchmark, we use the transfer learning performance on Pascal VOC semantic segmentation task given all Taskonomy models as sources.

We also investigate if the images selected to compute DDS have any effect on Spearman's correlation with transfer learning. For this purpose, we select images from NYUv2, Taskonomy, and Pascal VOC dataset and evaluate the proposed methods on both, Taskonomy and Pascal VOC benchmark. We further compute the variance performing bootstrap by randomly sampling 200 images from the same dataset 100 times to compute similarity. The bootstrap sampling generates a bootstrap distribution of correlation between transfer performance and similarity measures, which allows measuring the variance in Spearman's correlation with transfer performance when selecting different images from the same dataset. 

\begin{figure}[t!]

\begin{center}
   \includegraphics[width=1\linewidth]{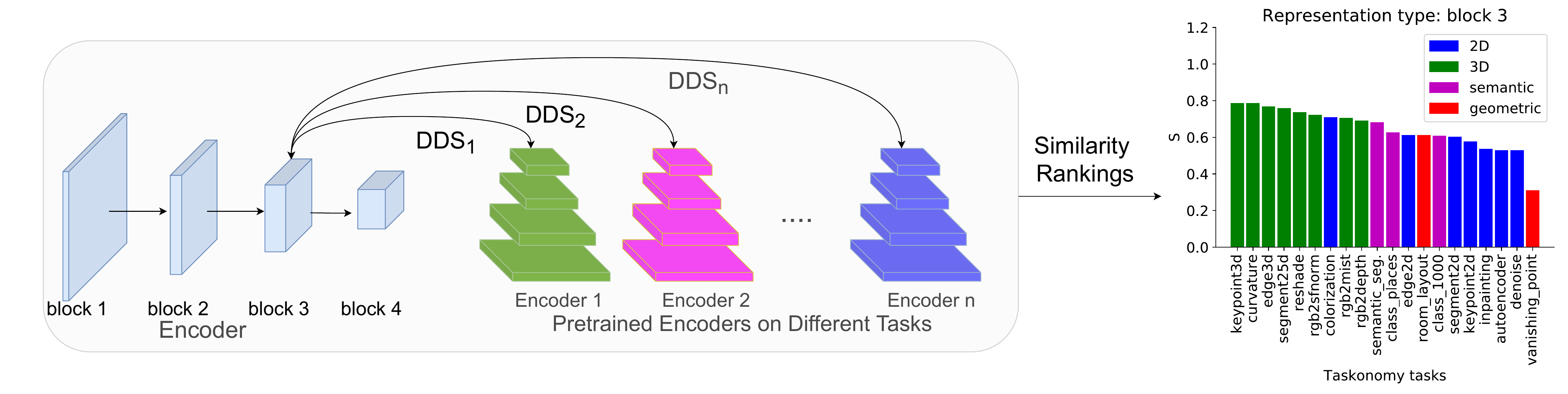}
\end{center}
   \caption{\emph{DNN Layer Selection.} Given a pre-trained encoder and a set of pre-trained models trained on diverse tasks, we can assess the representation type at different depth of the network by comparing the similarity between features at a given depth and pre-trained models.}%
\label{figure3}
\end{figure}

\subsection{Does DDS find best layer representation within a model to transfer from?}
In previous works \cite{zamir2018taskonomy,Dwivedi_RSA_19,song2019deep}, a major focus was to select a model to initialize from. However, once the model is selected as an encoder for initialization, the new layers of decoder usually branch out from the last layer of the pre-trained encoder. Such an approach is based on the \emph{a priori} assumption that, for any new task, the output from the last layer of the pre-trained encoder is the best representation for transfer learning. We argue that this is task-type dependent. For instance, it has been shown that earlier layers of DNNs trained on ImageNet object recognition learn low-level visual features while deeper layers learn high-level categorical features~\cite{olah2017feature}. Therefore, one would expect for low-level visual task, the representation in earlier layers of DNN might be better for transfer learning. Based on this intuition, we investigate if layers at different depths of the network are better suited to transfer to different types of tasks.
 We compute DDS of a pre-trained model at different depths with Taskonomy models to assess representation types at different depths (Fig. \ref{figure3}). To validate it, we select 3 task types (2D, 3D, and semantic) from NYUv2 and Pascal VOC dataset and perform transfer learning by attaching the decoder to different encoder layers.

\section{Experimental Setup}
We implemented the DDS general framework in python\footnote{Code available at https://github.com/cvai-repo/duality-diagram-similarity}, in which new  parameters and functions ($\mathbf{Q}$, $\mathbf{D}$, $f$, $g$) can be  incorporated in the future. Below, we first provide details of datasets and models used for comparing the DDS combinations for model selection. Then, we describe the datasets and models used for layer selection from a pre-trained encoder.

\subsection{Dataset and models for Model Selection}
\paragraph{\bf Datasets.}
To compare different DDS combinations against the Taskonomy affinity and winrate matrix, we randomly select 200 images from the Taskonomy dataset. We use 200 images based on an analysis that shows that the correlation of DDS with transfer learning performance saturates at around 200 images (see Supplementary S3). 
To perform the bootstrap based comparison on a new semantic segmentation task on the Pascal VOC dataset, we randomly select 5000 images from Taskonomy, 5000 images from Pascal VOC, and all (1449) images from NYUv2 dataset. 

\paragraph{\bf Models.}
We use the selected 200 images to generate features from the last layer of the encoder of 17 models trained on 17 different tasks on the Taskonomy dataset. The Taskonomy models have an encoder/decoder architecture. The encoder architecture for all the tasks is a fully convolutional Resnet-50 without pooling to preserve high-resolution feature maps. The decoder architecture varies depending on the task. The models were trained on different tasks independently using the same input images but different labels corresponding to different tasks. For comparing two models, we use the features of the last layer of the encoder following~\cite{zamir2018taskonomy,Dwivedi_RSA_19}. The Pascal VOC semantic segmentation model that we use also has the same encoder architecture as Taskonomy models, and the decoder is based on the spatial pyramid pooling module, which is suitable for semantic segmentation tasks~\cite{DeepLabV3}. For comparison with the Pascal VOC model, we use the features of the last layer of the encoder of 17 Taskonomy models and the one Pascal VOC semantic segmentation model trained from scratch. We also report comparison with a small Pascal VOC model from \cite{Dwivedi_RSA_19} in Supplementary S4 to show that model selection can be performed even using small models. 

\subsection{Dataset and models for layer selection}
\paragraph{\bf Datasets.}
To validate whether the proposed layer selection using similarity measures reflects transferability, we perform training on different datasets and tasks by branching the decoders from different layers of the encoder. Specifically, we evaluate on 3 tasks (Edge Detection, Surface Normal Prediction and Semantics Segmentation) on Pascal VOC~\cite{PascalVOC} dataset, and 3 tasks (Edge Detection, Depth Prediction and Semantic Segmentation) on NYUv2~\cite{NYUV2} dataset. Following Zamir \etal~\cite{zamir2018taskonomy}, we use Canny Edge Detector~\cite{canny1987computational} to generate groundtruth edge maps while other labels were downloaded from Maninis \etal~\cite{maninis2019attentive}.

\paragraph{\bf Models.}  We describe the models' encoder and decoder. 
\\
\noindent\emph{Encoder:} We use a ResNet-50~\cite{ResNet} pre-trained on ImageNet~\cite{deng2009imagenet} as our encoder, which has four blocks, each of the block consist of several convolution layers with skip connections, followed by a pooling layer. The branching locations that we explore are after each of the four pooling layers. We also consider Resnet-50 pre-trained on Places~\cite{places365} using the same experimental set-up, and report the results in Supplementary S5.
\\
\noindent\emph{Decoder:} Following the success of DeepLabV3~\cite{DeepLabV3} model, we use their decoder architecture in all our experiments. Since the output channels of the ResNet-50 encoder varies at different branching locations, we stack the output feature maps to keep the number of parameters in the downstream constant. More specifically, the encoder outputs 256, 512, 1024, 2048 channels for location 1, 2, 3 and 4 respectively, we stack the output of early branchings multiple times (8$\times$ for location 1, 4$\times$ for location 2 and 2$\times$ for location 3) to achieve a constant 2048 output channels to input to the decoder.

\paragraph{\bf Training.}
 ImageNet~\cite{deng2009imagenet} pre-trained encoder is fine-tuned for the specific tasks, while the decoder is trained from scratch. In all the performed experiments, we use synchronized SGD with momentum of 0.9 and weight decay of 1e-4. The
initial learning rate was set to 0.001 and updated with the "poly" learning rate policy~\cite{DeepLabV3}. The total number of epochs for the training was set to 60 and 200, for Pascal VOC~\cite{PascalVOC} and NYUv2~\cite{NYUV2}, respectively as in Maninis \etal~\cite{maninis2019attentive}.

\section{Results}
In this section, we first report the comparison results of different similarity measures. After selecting the best similarity measure we apply it for identifying the representation type at different depth of the pre-trained encoder. Finally, we validate if the branching selection suggested using similarity measures gives the best transfer performance, by training models with different branching locations on NYUv2 and Pascal VOC datasets. 

\begin{table*}[!t]
\centering

{
\begin{tabular}{c|cccccc}
 \multirow{2}{*}{\diagbox{$\mathbf{Q}$,$\mathbf{D}$}{$f$}} & \multicolumn{3}{c}{kernels} &\multicolumn{3}{c}{distances} \\
                  &  linear & Laplacian                & RBF     & Pearson & euclidean               & cosine \\ \hline 
Identity              & 0.632   & 0.815                    & 0. 800  & 0.823   & 0.688                   & 0.742 \\
Z-score           & 0.842   & \textbf{\textcolor{blue}{0.860}} & \textbf{0.841}   & 0.856  & \textbf{0.850}                   & \textbf{\textcolor{green}{0.864}} \\
Batch norm        & 0.729   & 0.852                    & 0.840   & \textbf{\textcolor{brown}{0.857}}   & 0.807                   & 0.850\\
Instance norm     & \textbf{0.849}   & 0.835                    & 0.838   & 0.850   & 0.847                   & 0.850\\
Layer norm        & 0.823   & 0.806                    & 0.786   & 0.823   & 0.813                   & 0.823 \\
Group norm        & 0.829   & 0.813                    & 0.790   & 0.829   & 0.814                   & 0.829 \\
                         
\end{tabular}
}
\caption{\emph{Finding best DDS combination ($\mathbf{Q}$, $\mathbf{D}$, $f$ ,$g$). } We report the results of comparison with transferability for differents sets of $\mathbf{Q}$, $\mathbf{D}$ and $f$. Top 3 scores are shown in  \textcolor{green}{green}, \textcolor{blue}{blue}, \textcolor{brown}{brown}, respectively. Best $\mathbf{Q}$, $\mathbf{D}$ for each $f$ is shown in bold.}%
\label{table2}
\end{table*}

\subsection{Finding best DDS combination ($\mathbf{Q}$, $\mathbf{D}$, $f$ ,$g$) for transferability }

We perform a thorough analysis to investigate which combinations of ($\mathbf{Q}$, $\mathbf{D}$, $f$, and $g$) of the DDS lead to higher correlation with transferability rankings. We focus on how to assign weights to different feature dimensions using $\mathbf{Q}$, $\mathbf{D}$ and distance functions $f$ to compute the pairwise similarity between observations. In Table~\ref{table2}, we report results on the correlation with transferability rankings showing the effect of applying combination of $\mathbf{Q}$ and $\mathbf{D}$ instantiated as identity, z-score, batch/instance/group/layer normalization, and using different distance/kernel function as $f$. For $g$ we use Pearson's correlation on unbiased centered dissimilarity matrices because it consistently showed a higher correlation with transfer learning performance (Supplementary Section S6). We observed a similar trend in results using Spearman's correlation for $g$ (Supplementary Section S7). 

In Table~\ref{table2} we report the mean correlation of all the columns of the Taskonomy winrate matrix with the corresponding columns of a DDS based affinity matrix, which serve as the measure for computing how each of the similarity measures best predicts the transferability performance for each model. We first observe the results when $\mathbf{Q}$ and $\mathbf{D}$ are identity matrices. Laplacian and RBF kernels outperform linear kernel. For distance functions, Pearson outperforms euclidean and cosine. A possible reason for the better performance of Pearson's could be due to its invariance to translation and scale.
 
 We next observe the effect of normalization using appropriate $\mathbf{Q}$ and $\mathbf{D}$. We observe that the correlation with transferability rankings improves for all distance and kernel functions especially for low-performance distance and kernel functions. The gain in improvement is highest using z-scoring in most of the cases. A possible reason for overall performance improvement is that applying z-scoring reduces the bias in distance computation due to feature dimensions having high magnitude but low variance. Hence, for our next experiments, we choose z-scoring and select the top performing $f$: Laplacian and cosine.

\begin{table*}[!t]
\centering
{
\begin{tabular}{l|cc|c}

 Method  &  Affinity  & Winrate& Total time(s)\\ \hline 

 {Taskonomy Winrate\cite{zamir2018taskonomy}}       & {0.988}& {1} &  {$1.6\times10^7$}\\
 {Taskonomy affinity\cite{zamir2018taskonomy}}      & {1} & {0.988} &  {$1.6\times10^7$}\\ 
\hline
{saliency\cite{song2019deep}}       & {0.605}& {0.600}  &  {$3.2\times10^3$}\\ 
{DeepLIFT\cite{song2019deep}}       & {0.681}& {0.682}  &  {$3.3\times10^3$}\\
{$\epsilon$-LRP\cite{song2019deep}} & {0.682}& {0.682}  &  {$5.6\times10^3$}\\

\hline
 {RSA\cite{Dwivedi_RSA_19}}       & {0.777} & {0.767}   & {$78.2$}       \\
\hline
{DDS ($f=cosine$)}      & \textcolor{green}{0.862}  & \textcolor{green}{0.864}  & {$84.14$}\\

{DDS ($f=Laplacian$) }  & \textcolor{blue}{0.860}  & \textcolor{blue}{0.860}  & {$103.36$}\\
\end{tabular}
}
\caption{\emph{Correlation of DDS based affinity matrices with Taskonomy affinity and winrate matrix,  averaged for 17 Taskonomy tasks, and comparison to state-of-the-art.} Top 2 scores are shown in  \textcolor{green}{green}, and \textcolor{blue}{blue}  respectively. For this experiment, $\mathbf{Q}$ and $\mathbf{D}$ are selected to perform z-scoring, in all DDS tested frameworks.}%

\label{table3}
\end{table*}

\subsection{Comparison with state-of-the-art on Taskonomy}
We first compare the DDS based affinity matrices on the Taskonomy transferability benchmark. To quantify in terms of mean correlation across all the tasks, we report mean correlation with Taskonomy affinity and winrate matrix in Table~\ref{table3}. In Table~\ref{table3} (also Supplementary Section S8), we observe that all the proposed DDS based methods outperform the state-of-the-art methods~\cite{song2019deep,Dwivedi_RSA_19} by a large margin. DDS ($f=cosine$) improves \cite{Dwivedi_RSA_19} and \cite{song2019deep} by $10.9\%$ ($12.6\%$) and $26.3\%$ ($26.6\%$) on affinity (winrate), respectively.  We report the correlation of different DDS based rankings with the rankings based on winrate and task affinities for 17 Taskonomy tasks in Supplementary S9 and find that proposed DDS based methods outperform state-of-the-art methods for almost all the tasks. We also report comparison using PR curve following \cite{song2019deep} in Supplementary S10. 

To compare the efficiency of different methods with respect to bruteforce approach, we report the computational budget required for different methods. A single forward pass of Taskonomy models on Tesla V100 GPU takes 0.022 seconds. Thus, for 17 tasks and 200 feedforward passes for each task, the total time for feedforward pass is 74.8 sec. Hence, the DDS based methods are several orders of magnitude faster than bruteforce approach, used in the Taskonomy approach~\cite{zamir2018taskonomy}, that requires several GPU hours to perform transfer learning on all the models. The number reported in Table~\ref{table3} for Taskonomy was calculated by taking the fraction of the total transfer time (47,886 hours for 3000 transfers) for $17^2$ transfers used for comparison in this work. Further, the time for obtaining DDS based rankings takes only a few seconds on CPU and is an order of magnitude faster than attribution maps based methods. 

\subsection{Evaluating robustness on a new task and dataset}
In the evaluation benchmark that we proposed, which was used in the above reported experiments, we considered models that were trained using images from the Taskonomy dataset, and the images used to compute the DDS were also from the same dataset. To evaluate the robustness of DDS against a new task and images used to compute DDS, we consider a new task, namely Pascal VOC semantic segmentation, and use images from different datasets to compute DDS. To evaluate effect of selecting different images within the same dataset, we perform bootstrap to estimate the variance in correlation with transferability.

\begin{table*}[!t]
\centering
{
\begin{tabular}{l|c|c|c}

Method & Taskonomy & Pascal VOC & NYUv2 \\ \hline

 {DDS ($f=cosine$)} & { 0.525 $\pm0.057$}                      & {0.722 $\pm0.049$} & {0.518 $\pm0.034$}\\

 {DDS ($f=Laplacian$)}& \textcolor{green}{0.5779 $\pm0.050$}    & \textcolor{green}{0.765 $\pm0.038$}  & \textcolor{green}{0.521  $\pm0.029$}\\
  
\end{tabular}
}

\caption{\emph{DDS correlation with transfer learning for Pascal VOC Semantic Segmentation.} Here each row represents a particular distance/kernel function as $f$, and each column represents a dataset. The values in the table are bootstrap mean correlation and standard deviation of a particular similarity measure computed using the image from a particular dataset. Top score is shown in  \textcolor{green}{green}.}%
\label{table4}
\end{table*}

In Table~\ref{table4}, we report the bootstrap mean and standard deviation of correlation of different similarity measures with transfer learning performance on the Pascal VOC semantic segmentation task. We observe that the similarity measures show a high correlation (\textgreater 0.70 for $f=cosine$ and \textgreater 0.75 for $f=Laplacian$) when using images from Pascal VOC, but low correlation when using images from another dataset (Taskonomy and NYUv2). We also observed a similar trend in Taskonomy benchmark (Supplementary Section S11). Thus, the similarity measure is effective when using images from the same distribution as the training images for the model of the new task.  We believe that using images from the same data distribution in DDS as the ones used to train the model on the new task for selecting the best initialization is important because the model in the new task is trained using data sampled from this distribution. Since high correlation of DDS ($f=Laplacian$) with transferability is obtained in all the investigated scenarios using the images from the dataset of the new task that we want to transfer to, we argue that this is the most suitable choice for estimating transferability as compared to other similarity measures and set-ups.

\subsection{Finding representation type at different depth of a model}
In the previous experiments, we demonstrated  DDS ability to select models for transfer learning to a new task, given a set of source models. Here,  we  use DDS to interpret the representation type at different depths of the model, which would allow us to select which model layer to transfer from for a given type of task. For this purpose, we generate the features of the last layer of the encoder of 20 Taskonomy models to get the representation of each task type. We then compute the DDS ($f=Laplacian$) of Taskonomy features with each output block of the pre-trained ImageNet model. We use images from the same data distribution (Taskonomy) as used in the trained models to reveal the correlation with each layer and task type, as suggested in the previous experiment.  

\begin{figure*}[t!]

\begin{center}
   \includegraphics[width=1\linewidth]{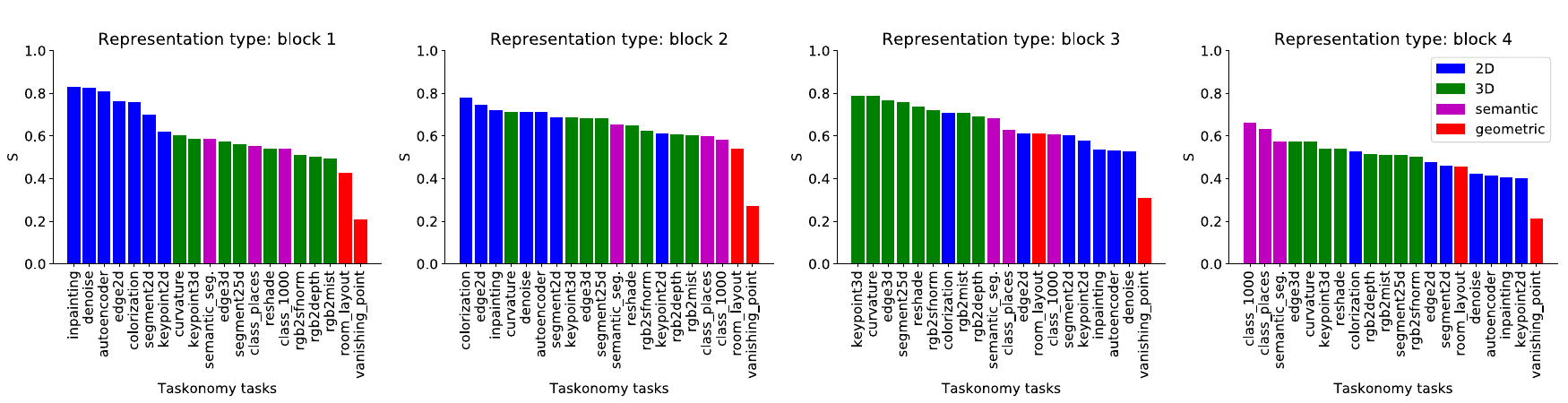}
\end{center}
   \caption{\emph{Block selection using DDS} on pre-trained encoder on Imagenet, and with DNNs trained on Taskonomy dataset on different tasks.}%
\label{figure5}
\end{figure*}

As shown in  Fig. \ref{figure5}, we observe that the representation of block 1 is more similar to 2D models, block 3 is more similar to 3D and block 4 to semantic models. These results suggest that the representation of block 1 is better suited to transfer for 2D tasks, block 3 for 3D tasks and block 4 for semantic tasks. There is no clear preference for block 2. We observe a similar pattern with the pre-trained Places model (see Supplementary S5).  
\subsection{Does DDS predict best branching location from an encoder?}
Here we report the results of transfer learning performances of 4 tasks: surface normal prediction on Pascal VOC~\cite{PascalVOC}, Depth Prediction on NYU Depth V2~\cite{NYUV2}, and edge detection and semantic segmentation on both datasets. These 4 tasks cover the 3 task clusters we observed in the previous section. The results are shown on Table~\ref{table5}. We report the qualitative comparison of different block outputs in Supplementary S5. We observe that branching out from block 3 gives the best performance on depth and surface normal, branching out from block 1 provides the best result on edge detection, and branching out from block 4 is best for semantic segmentation. The transfer learning results are consistent with the similarity results in Fig.~\ref{figure5}, which suggests that DDS ($f=Laplacian$) is a robust method for encoder block selection for different tasks.

\begin{table}[!t]
\centering
{
\begin{tabular}{c|c|c|c|c|c|c}
\hline \hline
                        & \multicolumn{3}{c|}{Pascal VOC}                                                                                                                                                   & \multicolumn{3}{c}{NYUv2}                                                                                                                                                    \\ \cline{2-7} 
\multirow{-2}{*}{\diagbox{Block}{Task}} & \begin{tabular}[c]{@{}c@{}}Edge\\ (MAE)\end{tabular} & \begin{tabular}[c]{@{}c@{}}Normals\\ (mDEG\_DIFF)\end{tabular} & \begin{tabular}[c]{@{}c@{}}Semantic\\ (mIOU)\end{tabular} & \begin{tabular}[c]{@{}c@{}}Edge\\ (MAE)\end{tabular} & \begin{tabular}[c]{@{}c@{}}Depth\\ (log RMSE)\end{tabular} & \begin{tabular}[c]{@{}c@{}}Semantic\\ (mIOU)\end{tabular} \\ \hline
1                       & {\color[HTML]{000000} \textbf{0.658}}                & 18.09                                                          & {\color[HTML]{000000} 0.257}                              & \textbf{0.823}                                       & 0.322                                                      & 0.124                                                     \\
2                       & 0.686                                                & 15.59                                                          & 0.392                                                     & 0.857                                                & 0.290                                                      & 0.165                                                     \\
3                       & 0.918                                                & \textbf{14.39}                                                 & 0.627                                                     & 1.297                                                & \textbf{0.207}                                             & 0.219                                                     \\
4                       & 0.900                                                & 15.11                                                          & \textbf{0.670}                                            & 1.283                                                & 0.208                                                      & \textbf{0.285}               \\ \hline \hline                            
\end{tabular}
}
\caption{\emph{Transfer learning performance of branching ImageNet pre-trained encoder on different tasks on Pascal VOC and NYUv2.} Results show that branching out from block 1, 3, 4 of the encoder have better performances on edge, normals (depth) and semantic tasks, respectively. This is consistent with the diagram similarity in Fig. \ref{figure5}.}%
\label{table5}
\end{table}

\section{Conclusion}

In this work, we investigated duality diagram similarity as a general framework to select model initialization for transfer learning. We found that after taking into account the weighing of feature dimension, DDS (for all distance functions) show a high correlation (\textgreater 0.84) with transfer learning performance.  We demonstrated on Taskonomy models that the DDS ($f=Laplacian$,$f=cosine$) shows $10\%$ improvement in correlation with transfer learning performance over state-of-the-art methods. DDS ($f=Laplacian$) is highly efficient and robust to novel tasks to create a duality diagram. We further show the  DDS ($f=Laplacian$) effectiveness in layer selection within a model to transfer from.

\subsubsection*{Acknowledgments.} G.R. thanks
the support of the Alfons and Gertrud Kassel Foundation. R.M.C. is supported by DFG grants (CI241/1-1, CI241/3-1) and the ERC Starting Grant (ERC-2018-StG 803370).

\bibliographystyle{splncs04}
\bibliography{egbib}
\renewcommand{\thesection}{S\arabic{section}} 
\renewcommand{\thefigure}{S\arabic{figure}} 
\renewcommand{\thetable}{S\arabic{table}}

\title{Supplementary Material\\Duality Diagram Similarity: \\ a generic framework for initialization selection \\ in task transfer learning} %

\titlerunning{Duality Diagram Similarity for Transfer Learning}

\author{Kshitij  Dwivedi\inst{1,3}\orcidID{0000-0001-6442-7140} \and
Jiahui Huang\inst{2}\orcidID{0000-0002-0389-1721} \and
Radoslaw Martin Cichy\inst{3}\orcidID{0000-0003-4190-6071} \and \\
Gemma Roig\inst{1}\orcidID{0000-0002-6439-8076} }

\authorrunning{Dwivedi et al.}
\institute{Department of Computer Science,  Goethe University Frankfurt, Germany \\
\email{kshitijdwivedi93@gmail.com, roig@cs.uni-frankfurt.de }\and
ISTD, Singapore University of Technology and Design, Singapore
\email{jiahui\_huang@sutd.edu.sg }\\ \and
Department of Education and Psychology, Free University Berlin, Germany\\
\email{rmcichy@zedat.fu-berlin.de}}
\maketitle
\noindent We provide the following items in the supplementary material, which complement the results reported in the main paper:
\begin{itemize}
\setlength\itemsep{0.25em}
\item[S1] RSA and CKA as a special case of duality diagram similarity (DDS).
\item[S2] Different normalizations in DDS Framework.
  \item[S3] Results on DDS's dependence on number of images.
  \item[S4] Results on model selection using coarse task representations.
  \item[S5] Quantitative and qualitative results of layer selection using a\\ ImageNet/Places365 pre-trained encoder.
  \item[S6] Effect of unbiased centering.
  \item[S7] Results with Spearman's correlation as $g$.
  \item[S8] DDS Results on Taskonomy and Pascal VOC for all distance/kernels as $f$.
  \item[S9] DDS Results for 17 Taskonomy tasks.
  \item[S10] Precision and Recall curves for DDS.
  \item[S11] DDS dependences on image dataset choice.

\end{itemize}
\section{RSA and CKA as special cases of duality diagram similarity (DDS)}

The duality diagram of a matrix $\mathbf{X} \in \mathbb{R}^{n\times d_1}$ can be calculated by the product of $\mathbf{Q_X}$, $\mathbf{X}$ and $\mathbf{D}$, where $\mathbf{Q} \in \mathbb{R}^{d_1\times d_1}$ is a matrix that quantifies dependencies between the individual feature dimensions, and $\mathbf{D} \in \mathbb{R}^{n\times n}$ is a matrix that assigns weights on the observations.

Let $\mathbf{\hat{X}}$ and $\mathbf{\hat{Y}}$ be the duality diagrams obtained from two different models (layers), the duality diagram similarity (DDS) between those two can be calculated by first computing pairwise distance matrices, $\mathbf{M_X}$, $\mathbf{M_Y}$, using a distance function, $f$, then use another function, $g$, to compare $\mathbf{M_X}$ and $\mathbf{M_Y}$ to obtain the final similarity score, $\mathbf{S}$. The formulation of DDS can be written as:
\begin{equation}
\mathbf{S} = g\bigg(f\Big(\mathbf{DXQ_X}\Big), f\Big(\mathbf{DYQ_Y}\Big)\bigg)
\label{eq:DDS}
\end{equation}

\subsubsection{RSA as DDS.} To compute RSA, one needs to obtain for each model (layer) the Representation Dissimilarity Matrices (RDMs), which is populated by computing a dissimilarity score $1-\rho$, where $\rho$ is the Pearson's correlation coefficient between each pair of images (observations). Once the RDMs for each model (layer) is computed, then Spearman's correlation of the upper triangular part of the 2 RDMs is used to compute the final similarity score between the two RDMs. Here, one can  observe the connection between RSA and DDS. In Equation~\ref{eq:DDS}, RDMs are the above-mentioned pairwise distance matrices, $\mathbf{M_X}$ and $\mathbf{M_Y}$, the distance function $f$ used in RSA is the dissimilarity score $1-\mathbf{\rho}$.  If no normalization is used, matrix $\mathbf{D}$ and matrix $\mathbf{Q}$ are  both identity matrices, $\mathbf{I}$ (ones in the diagonal and the rest of the elements in the matrix are zeros). In \cite{Dwivedi_RSA_19}, they use a centering matrix $\mathbf{C}$ ($\mathbf{C} = \mathbf{I_n}-\frac{1}{n}\mathbf{1_n}$, where $\mathbf{1}$ is the $n\times n$ matrix of all ones)   as $\mathbf{D}$ in the formulation of the duality diagram, and $\mathbf{Q}$ as the identity matrix. The final similarity score, $g$,  used in RSA is the Spearman's correlation between lower/upper triangular part of the two RDMs. Finally, RSA as a special case of DDS can be written as:
\begin{equation}
\mathbf{S} = r_s^t\bigg(1-\rho \Big(\mathbf{CXI}\Big), 1-\rho\Big(\mathbf{CYI}\Big)\bigg)
\end{equation}
where $r_s^t$ denotes the Spearman's correlation of the upper traingular part of the two input matrices, the dissimilarity score $1-\rho$ is computed with the Pearson's correlation, $\mathbf{C}$ is the centering matrix for $\mathbf{X}$ and $\mathbf{Y}$, and $\mathbf{I}$ is the identity matrix.%

\subsubsection{CKA as DDS.} The formulation of CKA~\cite{kornblith2019better} can be written as:
\begin{equation}
\mathbf{\text{CKA}\Big(X, Y\Big) = \text{tr}\Big(KHLH\Big)/\sqrt{\text{tr}\Big(KHKH\Big)\text{tr}\Big(LHLH\Big)}},
\end{equation}
in which $\mathbf{K}$ and $\mathbf{L}$ are the output matrices after applying either the RBF or linear kernel (kernel function $k$) on data matrices $\mathbf{X}$ and $\mathbf{Y}$, respectively. Mathematically, $\mathbf{K}$ and $\mathbf{L}$ can be expressed as , $\mathbf{K_{ij}} = k(\mathbf{x}_i, \mathbf{x}_j)=\exp(-{\gamma}_{2}||\textbf{x}_i - \textbf{x}_j||^2)$, $\mathbf{L_{ij}} = k(\mathbf{y}_i, \mathbf{y}_j)=\exp(-{\gamma}_{2}||\textbf{y}_i - \textbf{y}_j||^2)$ for RBF, and, $\mathbf{K_{ij}} = k(\mathbf{x}_i, \mathbf{x}_j) = \mathbf{x}_i^{T} \mathbf{x}_j$, $\mathbf{L_{ij}} = k(\mathbf{y}_i, \mathbf{y}_j) = \mathbf{y}_i^{T} \mathbf{y}_j$  for the linear kernel. Here $\mathbf{x}_i$ and $\mathbf{y}_i$ denote the $i^{th}$ column of $\mathbf{X}$ and $\mathbf{Y}$ respectively, $\mathbf{H}$ is the centering matrix($\mathbf{H} = \mathbf{I_n}-\frac{1}{n}\mathbf{1_n}$, where $\mathbf{1}$ is the $n\times n$ matrix of all ones) , and $T$ denotes the transpose. 
To obtain the equivalent formulation in DDS, in the following equations, we substitute $\mathbf{KH}$ and $\mathbf{LH}$ with $\mathbf{\hat{K}}$ and $\mathbf{\hat{L}}$ for simplification. Since $\mathbf{\hat{K}_{ij}} = \mathbf{\hat{K}_{ji}}$, we can get $\mathbf{K^T = K}$, similarly, $\mathbf{L^T = L}$, thus the above equation can be written as:
\begin{equation}
\begin{split}
    \mathbf{\text{CKA}\Big(X, Y\Big)} & = \sum_{ij}\mathbf{\hat{K}_{ij}\hat{L}_{ij}/\sqrt{\sum_{ij}\hat{K}^2_{ij}\sum_{ij}\hat{L}^2_{ij}}},
\end{split}
\end{equation}
which corresponds to the cosine similarity between $\mathbf{\hat{K}}$ and $\mathbf{\hat{L}}$. Since $\mathbf{K_{ij}} = k(x_i, x_j)$ and $\mathbf{L_{ij}} = k(y_i, y_j)$, we can treat them as pairwise distance matrix $\mathbf{M_X}$, $\mathbf{M_Y}$, respectively, in the formulation of DDS. In both cases, $\mathbf{Q}$ and $\mathbf{D}$ are both identity matrix here, and the distance function $f$ is the linear or the RBF kernel. The final similarity function $g$ used here is the cosine distance combined with the multiplication of the input matrices with the centering matrix $\mathbf{H}$. From above, we can derive CKA as a special case of DDS, and it can be written as:
\begin{equation}
\mathbf{S} = cos\bigg(k\Big(\mathbf{IXI}\Big)\mathbf{H}, k\Big(\mathbf{IYI\Big)\mathbf{H}\bigg)}
\end{equation}
where $cos$ is the cosine distance, $k$ is either the linear or the RBF kernel, and $\mathbf{I}$ is the identity matrix.

\section{Different normalizations in DDS framework}
In Table \ref{tableNorms}, we show how different normalizations used in deep learning and z-scoring can be reformulated in the DDS framework. Let $\mathbf{X} \in \mathbb{R}^{n\times c\times h\times w}$ be the output feature map of a convolutional layer with number of channels $c$, height $h$, width $w$ for $n$ input images. By swapping axes and reshaping the feature map $\mathbf{X}$, all the normalizations investigated in this work can be described in Duality Diagram setup. It is crucial to note that after reshaping the feature map, $\mathbf{D}$ and $\mathbf{Q}$ no longer represent weighing image and feature dimensions.

\begin{table}[!h]
\centering
{
\begin{tabular}{c|c|c|c}
\hline \hline
\diagbox{Norm}{}& \begin{tabular}[c]{@{}c@{}}$\mathbf{D}$\end{tabular} & \begin{tabular}[c]{@{}c@{}}$\mathbf{X}$\end{tabular} & \begin{tabular}[c]{@{}c@{}}$\mathbf{Q}$\end{tabular} \\ \hline \hline
Z-score         & $\mathbf{I}_{n\times n}-\mathbf{1}_{n\times n}/n$                           & $\mathbf{X}_{n\times chw}$                         & $\mathbf{S}_{chw\times chw}$                                                     \\
Batch Norm      & $\mathbf{I}_{nhw\times nhw}-\mathbf{1}_{nhw\times nhw}/nhw$                             & $\mathbf{X}_{nhw\times c}$                          & $\mathbf{S}_{c\times c}$                                                     \\
Group Norm      & $\mathbf{I}_{\frac{c}{g}hw\times \frac{c}{g}hw}- \mathbf{1}_{\frac{c}{g}hw\times \frac{c}{g}hw}/\frac{c}{g}hw$                             & $\mathbf{X}_{\frac{c}{g}hw\times ng}$                          & $\mathbf{S}_{ng\times ng}$                                                                                                          \\ \hline \hline
\end{tabular}

}
\caption{\emph{Different normalizations in DDS framework.} Here $\mathbf{I}$ denotes an identity matrix, $\mathbf{1}$ denotes matrix filled with all $1$'s, $\mathbf{X}$ is the output feature map of a convolutional layer with number of channels $c$, height $h$, width $w$ for $n$ input images, $g$ is group size for group normalization, and $\mathbf{S}$ is a diagonal matrix with diagonal values equal to standard deviation of $\mathbf{X}$ calculated across its rows. For each normalization, $\mathbf{X}$ is reshaped as indicated in the table. Layer and Instance normalization can be described by setting the group size $g$ in Group norm to $1$ and $c$ respectively. }
\label{tableNorms}
\end{table}

\section{DDS's dependence on number of images}
\begin{figure}[t]

\begin{center}
   \includegraphics[width=0.85\linewidth]{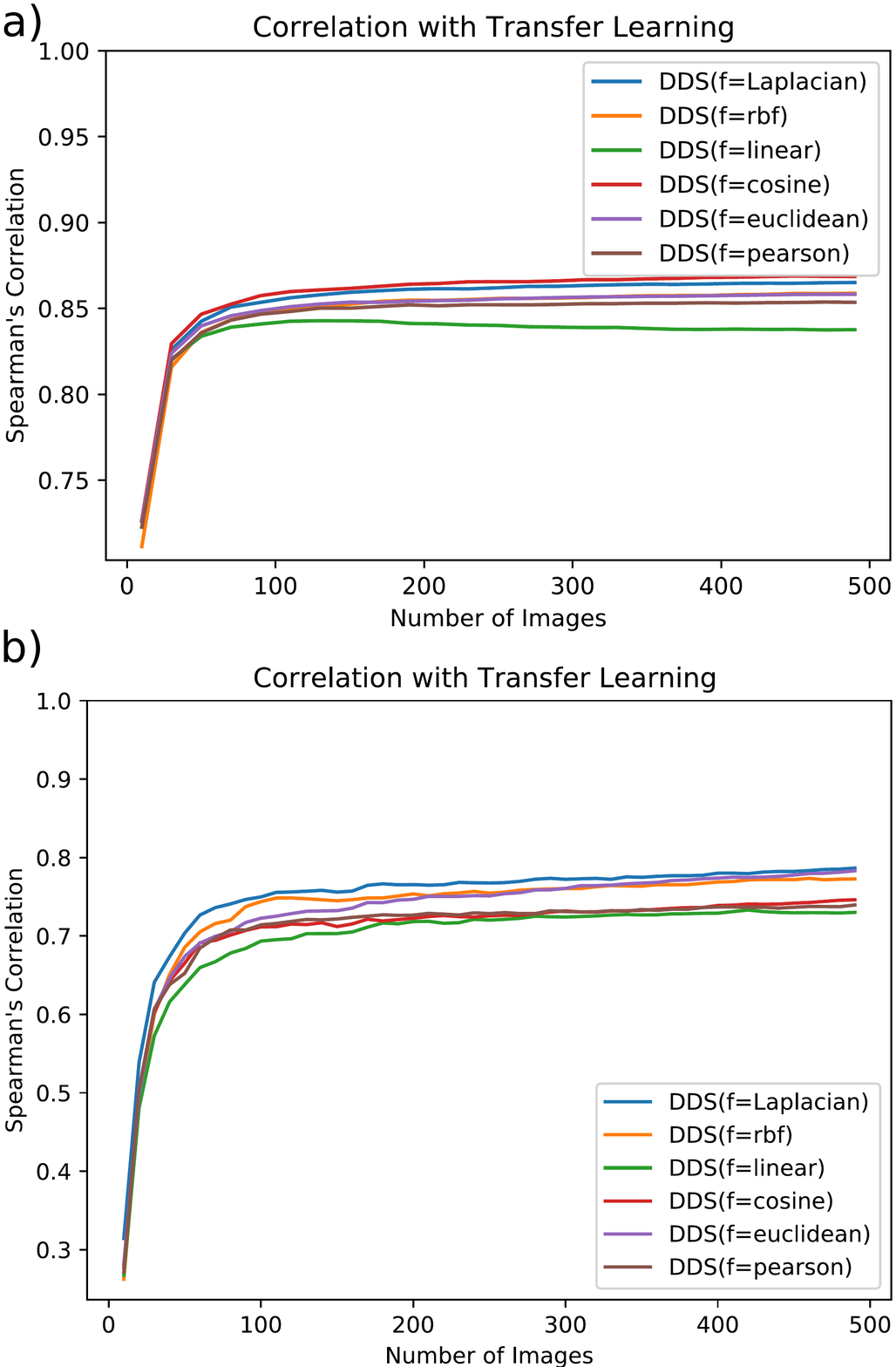}
\end{center}
   \caption{\emph{Spearman's correlation of DDS and transfer learning performance} on a) Taskonomy tasks, and b) Pascal VOC semantic segmentation task. The above plots shows  how Spearman's correlation of DDS with transfer learning varies with the number of images used to compute similarity using DDS with different distance/kernel functions as $f$. The images are randomly sampled from the Pascal VOC dataset.}
\label{Sfigure1}
\end{figure}

To calculate similarity between 2 Deep Neural Networks (DNNs) using DDS, we need to perform feedforward pass through both DNNs on a selected set of images. Here we analyse the impact of the number of images selected to compute the similarity measure. We varied the number of images from 10 to 500, in increments of 10, in a randomly selected set of Taskonomy images for Taskonomy benchmark and Pascal VOC images for Pascal VOC transfer learning benchmark, to calculate DDS.  We plotted the correlation with transfer learning on Taskonomy tasks and  on Pascal VOC semantic segmentation task in Figure~\ref{Sfigure1}a and Figure~\ref{Sfigure1}b respectively. We show the results for DDS with different $f$, namely Laplacian, RBF, linear, cosine, Euclidean and Pearson's correlation. We observe from the plots that correlation value with transfer learning saturates at around 200 images for all functions. For this reason, in all the experiments reported in the main paper  we use 200 images in the selected sets.

\section{Model selection using a coarse task representation}
\begin{figure}[t!]

\begin{center}
   \includegraphics[width=0.85\linewidth]{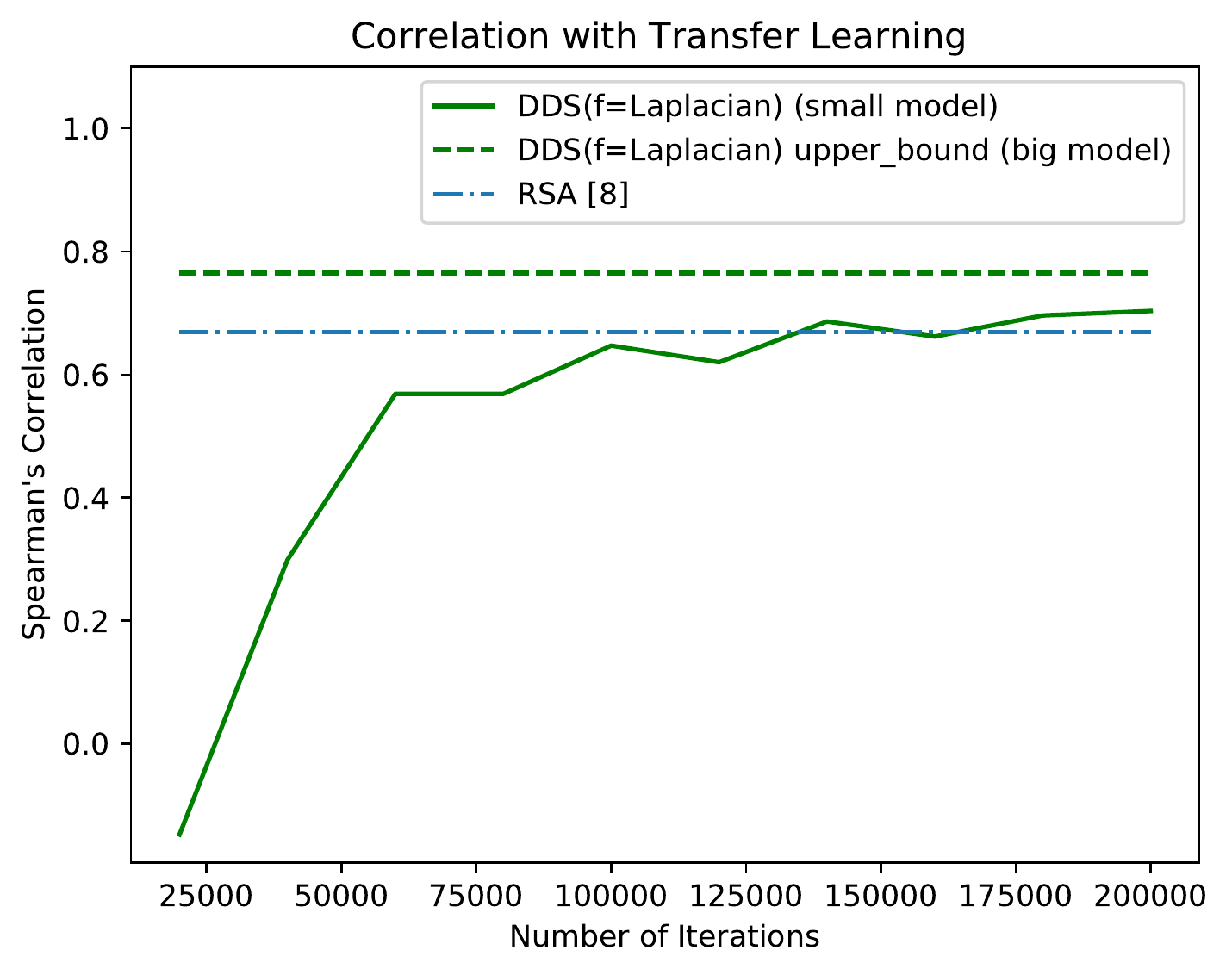}
\end{center}
   \caption{\emph{Correlation of DDS($f=Laplacian$) based rankings obtained using a small model trained on Pascal VOC semantic segmentation task  with the transfer learning performance.} We show how correlation varies with different stages of training. We further compare the results with the upper bound obtained by using the large trained Taskonomy type model on Pascal VOC semantic segmentation as the task representation and also a baseline RSA using the large model. }
\label{Sfigure22}
\end{figure}

Using task affinities as a method for source model selection, which is common  also in all related works \cite{Dwivedi_RSA_19,song2019deep,zamir2018taskonomy}, requires a pre-trained model on the new task itself to measure affinities. In \cite{Dwivedi_RSA_19}, it was proposed to train a small model on the new task, instead of a full large model, because it can be trained faster. The small models learn a coarse representation of the new task, and the task affinities to the source models can be compared faster.  We use the small model from \cite{Dwivedi_RSA_19}, and compare the correlation with transfer learning performance for DDS($f=Laplacian$) using small models and big models. We show the comparison in Figure~\ref{Sfigure22}, and we observe that correlation with transfer learning performance using small model is very close to the correlation using fully trained Taskonomy type model. Further, we observe that using DDS($f=Laplacian$) even with small model we outperform baseline RSA~\cite{Dwivedi_RSA_19} that uses fully trained Taskonomy type models. Overall, the above results suggest that even with a coarse representation obtained by training a small model on new task can assist in model selection using the similarity measures proposed in this work.

\section{Results of Layer selection(ImageNet/Places pre-trained encoder)}
\begin{figure*}[t!]

\begin{center}
   \includegraphics[width=1\linewidth]{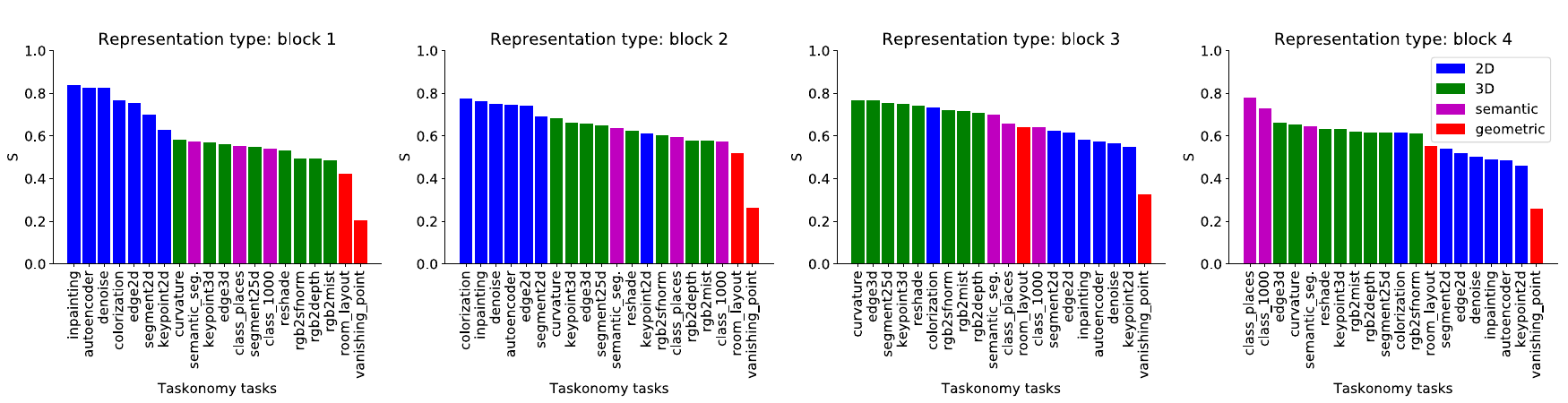}
\end{center}
   \caption{\emph{Block selection using DDS} on pre-trained encoder on Places, and with DNNs trained on Taskonomy dataset on different tasks.  }
\label{Sfigure33}
\end{figure*}

In addition to the experiments conducted with ImageNet pre-trained encoder, reported in the main paper, here we also provide results for an encoder pretrained on Places365~\cite{places365}. The representation type of different blocks of Places pre-trained model, as shown in Figure \ref{Sfigure33}, is similar to what we observed in Imagenet pre-trained model, reported in the main paper. From  Table~\ref{table2} and Table~\ref{table3}, we observe that our similarity measure successfully predicted best branching location for 5 out of 6 cases. Only exception is NYUv2 depth estimation task, where the transfer learning performance of block 3, selected by the method, is slightly lower than the best branching location (block 4). Overall from the above results combined with ImageNet results from the main text, we find that the proposed method reliably selects high performing branching locations to transfer to new tasks.

\begin{table}[!t]
\centering
{
\begin{tabular}{c|c|c|c}
\hline \hline
\diagbox{Block}{Task}& \begin{tabular}[c]{@{}c@{}}Edge\\ (MAE)\end{tabular} & \begin{tabular}[c]{@{}c@{}}Normals\\ (mDEG\_DIFF)\end{tabular} & \begin{tabular}[c]{@{}c@{}}Semantic\\ (mIOU)\end{tabular} \\ \hline \hline
1                         & \textbf{0.680}                                       & 17.89                                                          & 0.244                                                     \\
2                         & 0.777                                                & 15.62                                                          & 0.368                                                     \\
3                         & 1.012                                                & \textbf{14.35}                                                 & 0.532                                                     \\
4                         & 1.002                                                & 14.73                                                          & \textbf{0.616}                                            \\ \hline \hline
\end{tabular}

}
\caption{\emph{Transfer learning performance of branching Places pre-trained encoder on 3 tasks on Pascal VOC dataset.} The results  indicate that branching out from block 1, 3, 4 of the encoder have better performances on edge, normals and semantic tasks, respectively.}
\label{table2}
\end{table}

\begin{table}[!t]
\centering

{
\begin{tabular}{c|c|c|c}
\hline \hline
\diagbox{Block}{Task} & \begin{tabular}[c]{@{}c@{}}Edge\\ (MAE)\end{tabular} & \begin{tabular}[c]{@{}c@{}}Depth\\ (log RMSE)\end{tabular} & \begin{tabular}[c]{@{}c@{}}Semantic\\ (mIOU)\end{tabular} \\ \hline\hline
1                         & \textbf{1.027}                                       & 0.320                                                     & 0.125                                                     \\
2                         & 1.188                                                & 0.286                                                     & 0.167                                                     \\
3                         & 1.183                                                & 0.223                                            & 0.216                                                     \\
4                         & 1.120                                                & \textbf{0.201}                                                     & \textbf{0.291}                                            \\ \hline\hline
\end{tabular}

}
\caption{\emph{Transfer learning performance of branching Places pre-trained encoder on 3 tasks on NYUv2 dataset.} The results are mostly consistent with branching location prediction based on DDS.}
\label{table3}
\end{table}

\begin{figure*}[t!]

\begin{center}
   \includegraphics[width=0.9\linewidth]{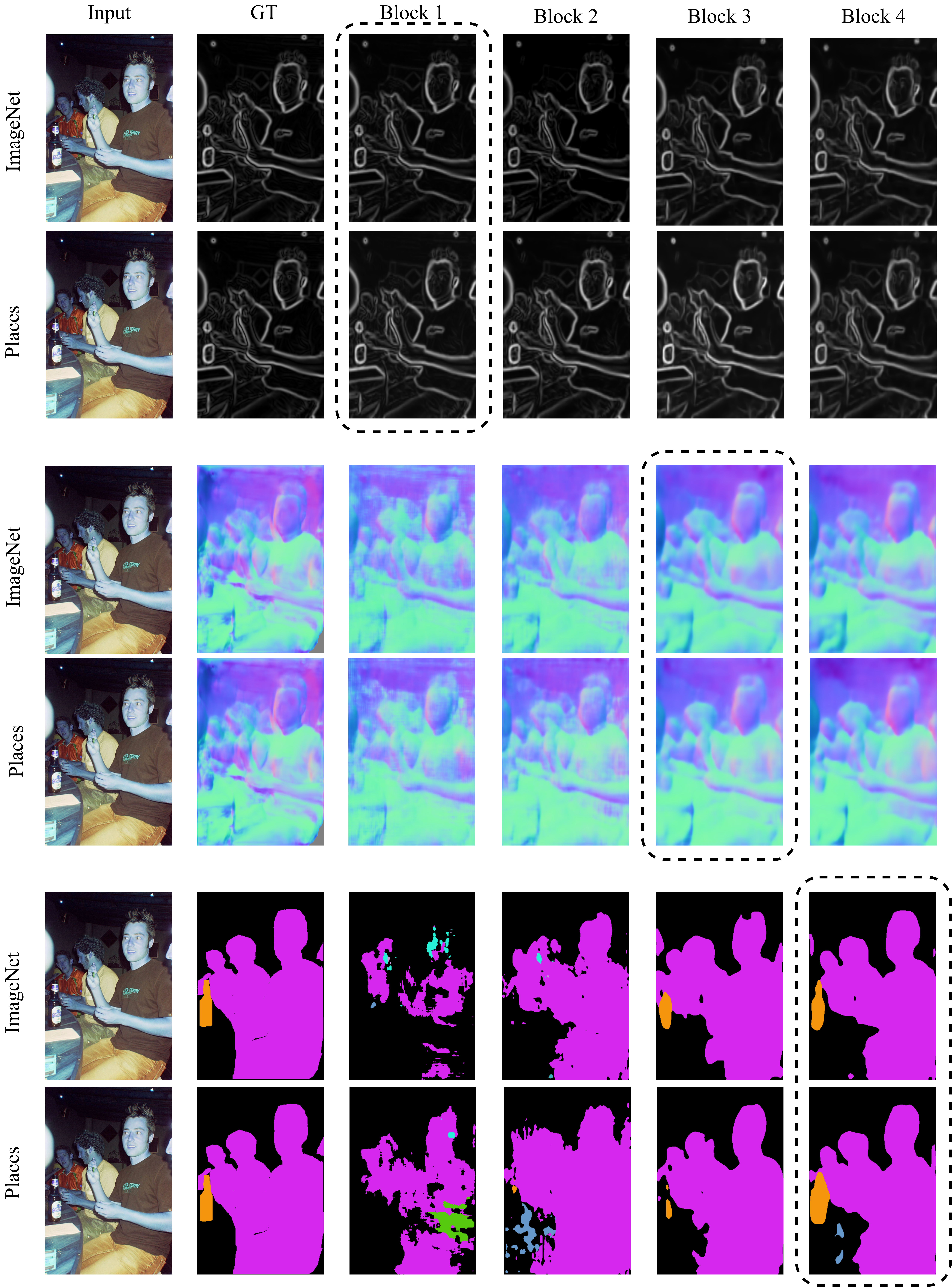}
\end{center}
   \caption{\emph{\textbf{Qualitative Results on Pascal VOC.} Branching results of all locations on three tasks are shown: Edge Detection, Surface Normal Prediction and Semantic Segmentation. For each task, ImageNet pre-trained encoder are shown on the upper row, while Places 365 pre-trained encoder are shown on the lower row. Best results are circled with dotted lines.}}
\label{Sfigure4}
\end{figure*}

\begin{figure*}[t!]

\begin{center}
   \includegraphics[width=1\linewidth]{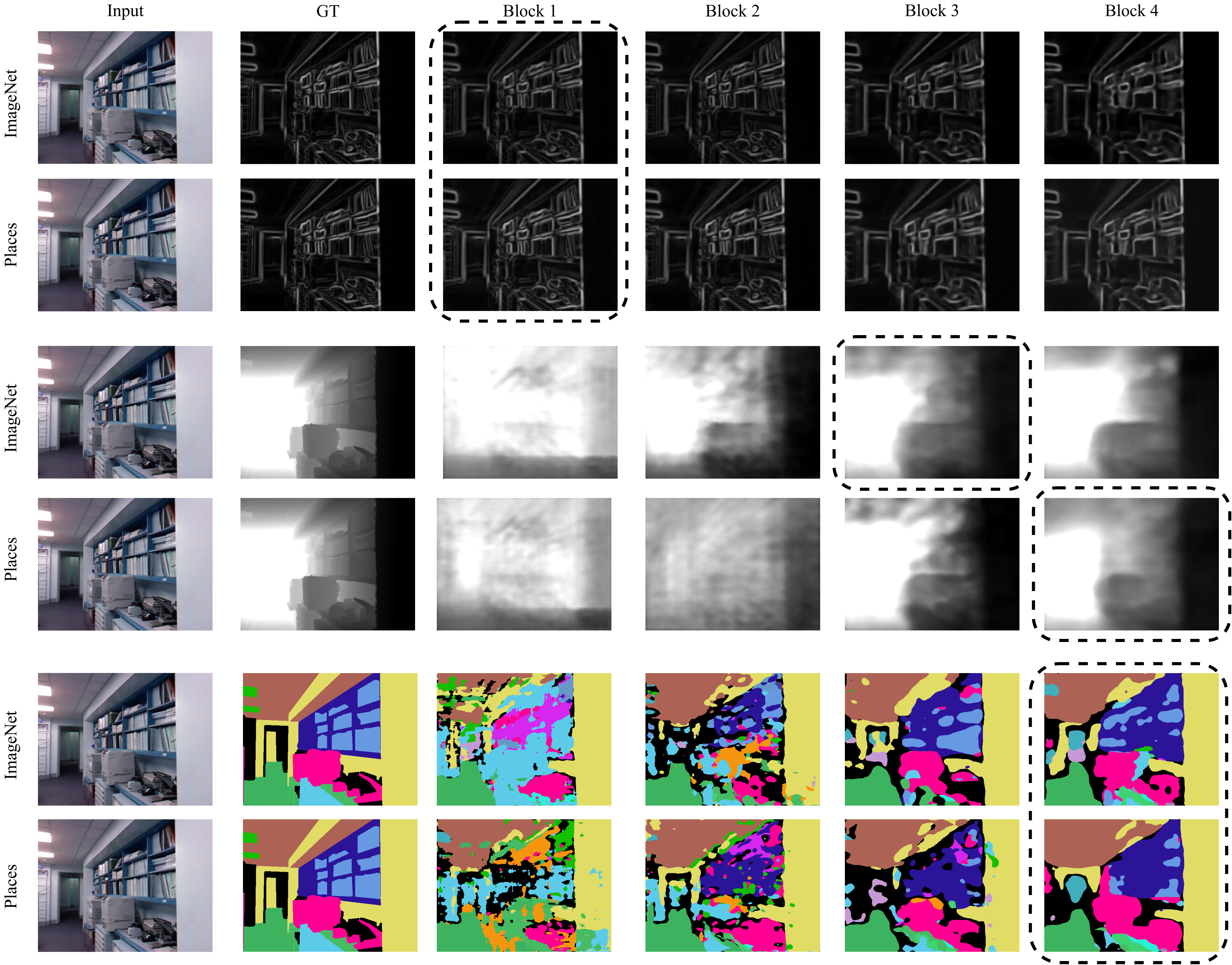}
\end{center}
   \caption{\emph{\textbf{Qualitative Results on NYUv2.} Branching results of all locations on three tasks are shown: Edge Detection, Depth Prediction and Semantic Segmentation. For each task, ImageNet pre-trained encoder are shown on the upper row, while Places 365 pre-trained encoder are shown on the lower row. Best results are circled with dotted lines.}}
\label{Sfigure5}
\end{figure*}

We show qualitative results on Pascal VOC\cite{PascalVOC} and NYUv2\cite{NYUV2} datasets in Figure~\ref{Sfigure4} and Figure~\ref{Sfigure5}. Here we illustrate branching results of 3 tasks: Edge Detection, Surface Normal (Depth) Prediction and Semantic Segmentation. For each task, ImageNet pre-trained encoder results are shown on the upper row, and Places 365 pre-trained encoder results are shown on the lower row. We observed some visual quality degradation in the results of non-optimal branching locations predicted by our similarity measures: Edge contours become blurry as the branching location goes deeper; semantic segmentation maps become closer to the ground truth at deeper layers.

\section{Effect of unbiased centering}
We report in Table \ref{tableC} the effect of applying unbiased centering (eq. 3.1 in \cite{szekely2014partial}) to $\mathbf{M_X}$ and $\mathbf{M_Y}$ on the correlation with transferability. We observe that for all  cases unbiased centering improves the correlation with transfer learning, and hence, in all the reported results in the main paper we used unbiased centering.

\begin{table*}[!t]
\centering

{
\begin{tabular}{c|cccccc}
 \multirow{2}{*}{\diagbox{Centering \\($\mathbf{M_X}$, $\mathbf{M_Y}$)}{$f$}} & \multicolumn{3}{c}{kernels} &\multicolumn{3}{c}{distances} \\\\
                  &  linear & Laplacian                & RBF     & Pearson & euclidean               & cosine \\ \hline 
No centering         & 0.818   & 0.691                    & 0.690  & 0.776   &0.613                  & 0.792 \\
Unbiased centering   & 0.842 & 0.860 & 0.841   & 0.856  & 0.850                   & 0.864\\
\end{tabular}
}
\caption{\emph{Effect of unbiased centering. } We report the results of comparison with transferability on Taskonomy transfer learning for with and without unbiased centering on pairwise (dis)similarity matrices $\mathbf{M_X}$ and $\mathbf{M_Y}$ }
\label{tableC}
\end{table*}

\section{Results with Spearman's correlation as $g$}

In the main text, we reported the results with $g$ as Pearson's correlation between unbiased centered (dis)similarity matrices $\mathbf{M_X}$ and $\mathbf{M_Y}$. Here, in Table~\ref{tableSpear}, we report 
 results when $g$ is the Spearman's correlation between upper/lower triangular part of unbiased centered (dis)similarity matrices $\mathbf{M_X}$ and $\mathbf{M_Y}$, as in \cite{Dwivedi_RSA_19}, on Taskonomy transfer learning benchmark. We observe that the results show similar trend with Spearman's correlation (improvement on applying z-scoring on $\mathbf{X}$,$\mathbf{Y}$) as using Pearson's correlation as $g$, shown in main text Table 2 .

\begin{table*}[!t]
\centering

{
\begin{tabular}{c|cccccc}
 \multirow{2}{*}{\diagbox{$\mathbf{Q}$}{$f$}} & \multicolumn{3}{c}{kernels} &\multicolumn{3}{c}{distances} \\
                  &  linear & Laplacian                & RBF     & Pearson & euclidean               & cosine \\ \hline 
Identity        & 0.778   & 0.828                    &  0.803  & 0.816   & 0.798                  & 0.803 \\
Z-score         & 0.858 & 0.864 & 0.846   & 0.844  & 0.862  & 0.860\\

\end{tabular}
}
\caption{\emph{Spearman's as $g$. } We report the results of comparison with transferability on Taskonomy transfer learning benchmark for with and without z-scoring when using Spearman's as $g$. }
\label{tableSpear}
\end{table*}

\section{DDS results for all distance/kernels as $f$ }

In Table 3 and Table 4 of main paper we  reported the best $f$ selected using the results in Table 2. Here we report the complete results for Table 3 and Table 4 with all investigated functions as $f$. Due to efficiency of our method it was possible to perform multiple bootstrap to calculate standard deviation in correlation with transfer learning. In the tables below (Table~\ref{tableTaskonomy} and Table~\ref{tablePascal}), we report bootstrap mean and standard deviation of correlation with transfer learning for Taskonomy tasks and Pascal VOC semantic segmentation task. We observe that DDS($f=Laplacian$) is the most robust (in Top 1,2 ) measure across both Taskonomy benchmark and Pascal VOC semantic segmentation tranfer learning.

\begin{table*}[!t]
\centering
{
\begin{tabular}{l|cc|c}

 Method  &  Affinity  & Winrate\\ \hline 
\hline
{DDS($f=pearson$)}      & {0.853 $\pm0.090$}  & {0.851 $\pm0.090$} \\
{DDS($f=euclidean$)}      & {0.852 $\pm0.076$}  & {0.855 $\pm0.079$} \\
{DDS ($f=cosine$)}      & \textcolor{green}{0.862 $\pm0.076$}  & \textcolor{green}{0.863 $\pm0.078$}  \\
{DDS($f=linear$)}      & {0.837 $\pm0.084$}  & {0.841 $\pm0.088$}  \\

{DDS ($f=Laplacian$) }  & \textcolor{green}{0.862 $\pm0.072$}  & \textcolor{blue}{0.861 $\pm0.072$} \\
{DDS($f=rbf$)}      & {0.854 $\pm0.086$}  & {0.854 $\pm0.088$}  \\
\end{tabular}
}
\caption{\emph{Correlation (Bootstrap mean $\pm standard dev$) of DDS based affinity matrices with Taskonomy affinity and winrate matrix,  averaged for 17 Taskonomy tasks.} Top 2 scores are shown in  \textcolor{green}{green}, and \textcolor{blue}{blue}  respectively. For this experiment, $\mathbf{Q}$ is set to z-scoring and $\mathbf{D}$ to the identity matrix, in all DDS tested frameworks.\vspace*{-0.6cm}}

\label{tableTaskonomy}
\end{table*}

\begin{table*}[!t]
\centering
{
\begin{tabular}{l|c|c|c}

Method & Taskonomy & Pascal VOC & NYUv2 \\ \hline 

 {DDS ($f=Pearson$)} & {0.534 $\pm0.063$}                     & {0.726 $\pm0.049$}                 & {0.505 $\pm0.033$}\\
 {DDS ($f=euclidean$)} & {0.534 $\pm0.055$}   & {0.746 $\pm0.051$}                 & {0.518 $\pm0.030$}\\
 {DDS ($f=cosine$)} & { 0.525 $\pm0.057$}                      & {0.722 $\pm0.049$} & {0.518 $\pm0.034$}\\
 {DDS ($f=linear$)} & {0.496 $\pm0.063$}                       & {0.718 $\pm0.062$}                 & {0.515 $\pm0.033$} \\

 {DDS ($f=Laplacian$)}& \textcolor{blue}{0.577 $\pm0.050$}    & \textcolor{green}{0.765 $\pm0.038$}  & \textcolor{blue}{0.521  $\pm0.029$}\\
  
  {DDS ($f=RBF$)}& \textcolor{green}{0.591 $\pm0.053$}          & \textcolor{blue}{0.753 $\pm0.051$}                 & \textcolor{green}{0.534 $\pm0.030$}\\
\end{tabular}
}\caption{\emph{DDS correlation with transfer learning for Pascal VOC Semantic Segmentation.} Here each row represents DDS with a particular distance/kernel function as $f$, and each column represents the dataset from which the images were selected to get similarity scores. The values in the table are bootstrap mean correlation and standard deviation of a particular similarity measure computed using the image from a particular dataset. Top score is shown in  \textcolor{green}{green}.\vspace*{-0.6cm} }
\label{tablePascal}
\end{table*}

\section{DDS similarity measure comparison for 17 Taskonomy tasks}

\begin{figure*}[t!]

\begin{center}
   \includegraphics[width=1\linewidth]{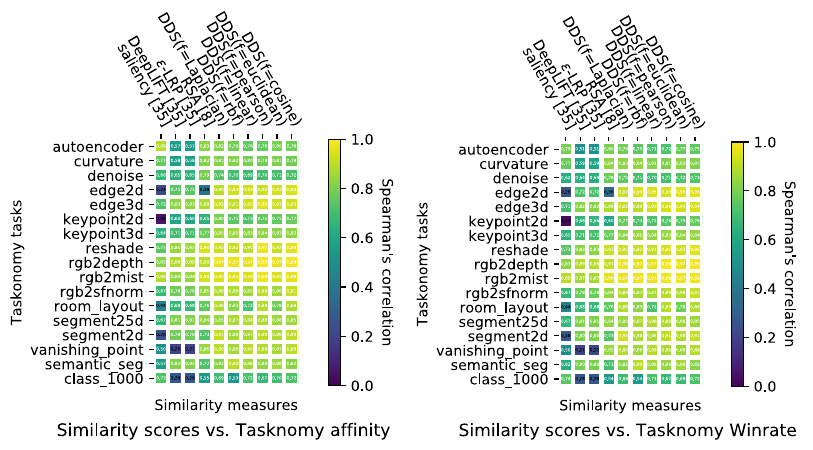}
\end{center}
   \caption{\emph{Similarity measures' comparison on Taskonomy Tasks.} Spearman's correlation of different similarity measure based rankings with transfer learning performance based rankings from Taskonomy affinity matrix (left), and Taskonomy winrate matrix (right) for 17 Taskonomy tasks as target. We show the results for 17 Taskonomy tasks (rows) for different similarity measures (columns). More yellow indicates higher the correlation, hence, is better.}
\label{Sfigure2}
\end{figure*}

In the main paper, we reported the mean correlation of similarity measures with transfer learning across 17 Taskonomy tasks. In Figure~\ref{Sfigure2}, we provide the detailed results on all tasks. We find that almost on all the tasks our proposed similarity measures outperform the state-of-the-art method  \cite{song2019deep,Dwivedi_RSA_19}.

\section{Precision and Recall curve for DDS}

\begin{figure}[t!]

\begin{center}
   \includegraphics[width=1\linewidth]{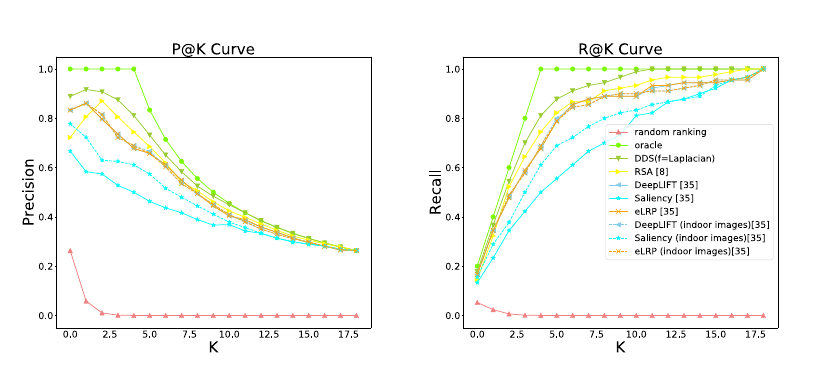}
\end{center}
   \caption{\emph{Precision and Recall Curve for comparing similarity measures.}  The x-axis in all the plots above refers to the number of source tasks used for calculating precision and recall value.   }
\label{Sfigure3}
\end{figure}

In the main text, we used correlation of similarity measure based source model rankings with transfer learning performance based rankings as our evaluation criteria. Song \etal~\cite{song2019deep} used precision and recall of selecting top-5 source tasks as the evaluation criteria. We use the evaluation code provided by \cite{song2019deep}, and we plot precision and recall curve for one of our most robust proposed method, DDS($f=Laplacian$), against state-of-the-art methods \cite{song2019deep,Dwivedi_RSA_19}, in Figure \ref{Sfigure3}. In Figure \ref{Sfigure3}, we plot results using 200 Taskonomy images for all the similarity measures that we compared. We further add the results of the methods from Song \etal~\cite{song2019deep} using 1000 images from indoor dataset used in Song \etal~\cite{song2019deep} that showed best performance in their paper. We observe from Precision and Recall plots in Figure \ref{Sfigure3} that DDS($f=Laplacian$) outperforms the state-of-the-art methods.

\section{DDS dependences on image dataset choice}
In this section, we investigate the effect of image dataset used to calculate Duality Diagrams. We report the results of DDS correlation with Taskonomy winrate in Table Table~\ref{tableTaskonomy_imsets}when images from Taskonomy, Pascal VOC, and NYUv2 were used to calculate Duality Diagrams. We observe a slight drop in DDS's correlation with Taskonomy winrate matrix when using images from Pascal VOC and NYUv2 dataset. 

These results are consistent with \cite{song2019deep} where they show that their method is robust to choice of images used to compute similarity between neural networks. In the aforementioned results, both source and target tasks were trained using the same training dataset, i.e. Taskonomy, and we believe that is the reason we, as well as \cite{song2019deep}, do not observe much difference.

However, when we compare transferability on Pascal VOC, source models are trained on Taskonomy dataset and target task is on Pascal VOC, which has significantly different statistics than Taskonomy. In this more challenging setting, we observe the impact of using images from different datasets, as reported in Section 6.3 in the main text.

\begin{table*}[!t]
\centering
{
\begin{tabular}{l|c|c|c}

Method & Taskonomy & Pascal VOC & NYUv2 \\ \hline 

 {DDS ($f=cosine$)} & { 0.864 }     & {0.818}                 & {0.822}\\

 {DDS ($f=Laplacian$)}& {0.860}     & {0.811}                 & {0.818}\\
  
\end{tabular}
}\caption{\emph{DDS correlation with transfer learning on Taskonomy Tasks.} Here each row represents DDS with a particular distance/kernel function as $f$, and each column represents the dataset from which the images were selected to obtains similarity scores. }
\label{tableTaskonomy_imsets}
\end{table*}

\end{document}